\documentclass[11pt,a4paper]{article}
\usepackage[hyperref]{acl2020}
\usepackage{times}
\usepackage{latexsym}
\usepackage{amsmath}
\usepackage{amssymb}
\usepackage{mathtools}
\usepackage{url}
\usepackage{dirtytalk}
\usepackage{enumitem}
\usepackage{graphicx}
\usepackage{todonotes}
\usepackage{xcolor}
\usepackage{booktabs}
\usepackage{xspace}
\usepackage{bbm}
\usepackage{adjustbox}
\usepackage{algorithm}
\usepackage{algpseudocode}
\usepackage{todonotes}

\usepackage[smallerops]{newtxmath}

\DeclareMathAlphabet{\mathcal}{OMS}{cmsy}{m}{n}
\DeclareMathAlphabet{\mathbb}{U}{msb}{m}{n}

\newcommand{\any}{\texttt{\textsc{entity}}\xspace}
\newcommand{\other}{\texttt{\textsc{other}}\xspace}

\newcommand{\pr}{^\prime}
\newcommand{\parent}[1]{\bar #1}
\newcommand{\children}{\operatorname{Ch}}
\newcommand{\sibling}{\operatorname{Sb}}
\newcommand{\level}{\operatorname{lev}}
\newcommand{\st}{<:}

\aclfinalcopy 

\title{Hierarchical Entity Typing via Multi-level Learning to Rank}

\author{
  Tongfei Chen ~~~~ Yunmo Chen ~~~~ Benjamin Van Durme\\
  Johns Hopkins University \\
  \texttt{\string{tongfei, ychen, vandurme\string}@cs.jhu.edu}
}

\date{}

\begin{document}
\maketitle

\begin{abstract}
 We propose a novel method for hierarchical entity classification that embraces ontological structure at both training and during prediction.  At training, our novel multi-level learning-to-rank loss compares positive types against negative siblings according to the type tree.  During prediction, we define a coarse-to-fine decoder that restricts viable candidates at each level of the ontology based on already predicted parent type(s). We achieve state-of-the-art across multiple datasets, particularly with respect to strict accuracy. 
\end{abstract}

\section{Introduction}
  Entity typing is the assignment of a semantic label to a span of text, where that span is usually a \emph{mention} of some entity in the real world.
  Named entity recognition (NER) is a canonical information extraction task, commonly considered a form of entity typing that assigns spans to one of a handful of types, such as  {\tt PER}, {\tt ORG}, {\tt GPE}, and so on.  Fine-grained entity typing (FET) seeks to classify spans into types according to more diverse, semantically richer ontologies \cite{LingW12,YosefBHSW12,GillickLGKH14,CorroAGW15,ChoiLCZ18}, and has begun to be used in downstream models for entity linking \cite{GuptaSR17,RaimanR18}.
  
\begin{figure}[t]
      \centering
      \includegraphics[width=0.9\linewidth]{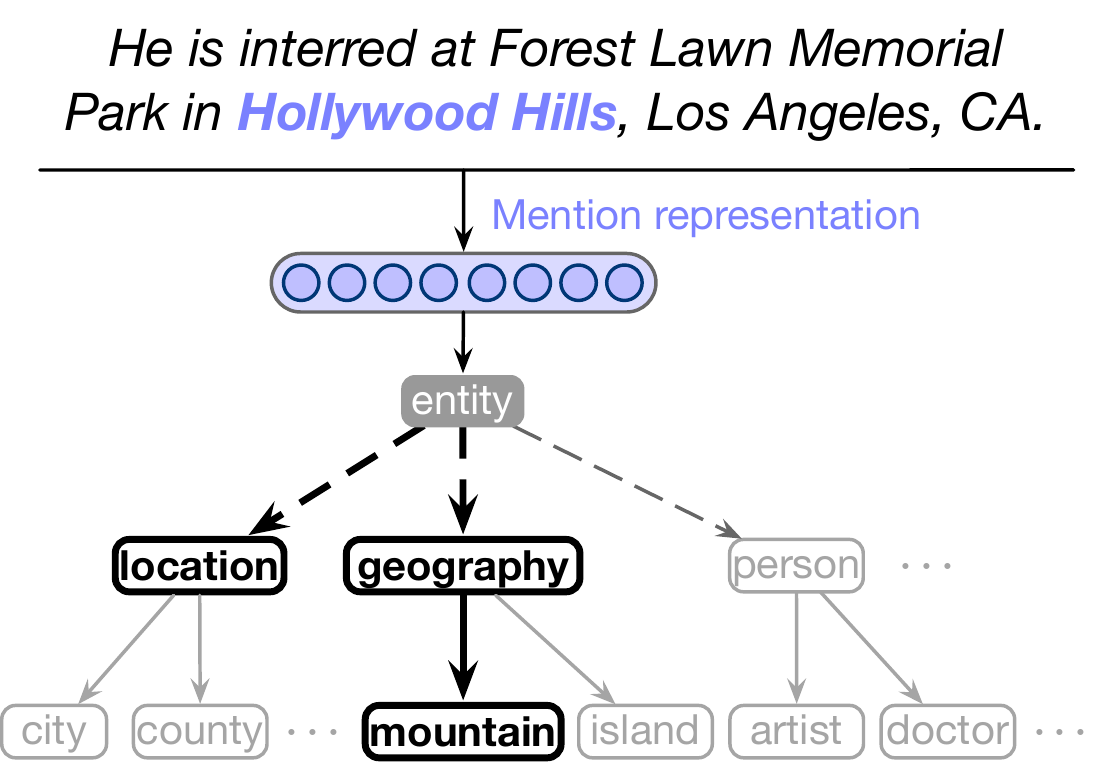}
      \caption{An example mention classified using the FIGER ontology. Positive types are highlighted.}
      \label{fig:figer-example}
  \end{figure}

  Consider the  example in \autoref{fig:figer-example} from the FET dataset, FIGER~\cite{LingW12}.
  The mention of interest, \emph{Hollywood Hills}, will be typed with the single label {\tt LOC} in traditional NER, but may be typed with a \emph{set} of types $\{${\tt /location}, {\tt /geography}, {\tt /geography/mountain}$\}$ under a fine-grained typing scheme.
  In these finer-grained typing schemes, types usually form a hierarchy: there are a set of coarse types that lies on the top level---these are similar to traditional NER types, e.g. {\tt /person}; additionally, there are finer types that are \emph{subtypes} of these top-level types, e.g. {\tt /person/artist} or {\tt /person/doctor}. 
  
  \begin{figure*}[t!]
      \centering
      \includegraphics[width=\linewidth]{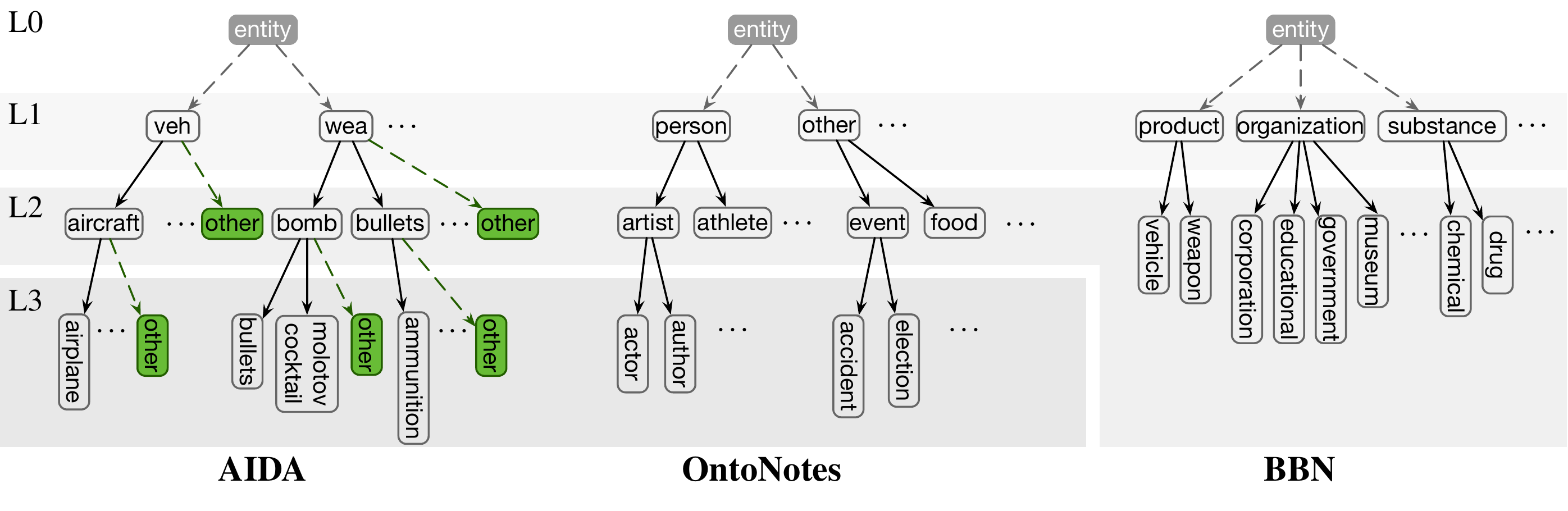}
      \caption{Various type ontologies. Different levels of the types are shown in different shades, from L0 to L3. The \any and \other special nodes are discussed in \autoref{sec:problem}.}
      \label{fig:ontologies}
  \end{figure*}
  
  Most prior  work concerning fine-grained entity typing has approached the problem as a \emph{multi-label classification} problem: given an entity mention together with its context, the classifier seeks to output a set of types, where each type is a node in the hierarchy. Approaches to FET include hand-crafted sparse features to various neural architectures \cite[\textit{inter alia}, see \autoref{sec:background}]{RenHQHJH16,ShimaokaSIR17,LinJ19}.

  Perhaps owing to the historical transition from ``flat'' NER types, there has been relatively little work in FET that exploits ontological \emph{tree structure}, where type labels satisfy the \emph{hierarchical property}: \emph{\textbf{a subtype is valid only if its parent supertype is also valid.}} We propose a novel method that takes the explicit ontology structure into account, by a \emph{multi-level learning to rank} approach that ranks the candidate types conditioned on the given entity mention. Intuitively, coarser types are easier whereas finer types are harder to classify: we capture this intuition by allowing distinct margins at each level of the ranking model. Coupled with a novel coarse-to-fine decoder that searches on the type hierarchy, our approach guarantees that predictions do not violate the hierarchical property, and achieves state-of-the-art results according to multiple measures across various commonly used datasets.

\section{Related Work}
\label{sec:background}

FET is usually studied as allowing for sentence-level context in making predictions, notably starting with \citet{LingW12} and \citet{GillickLGKH14}, where they created the commonly used FIGER and OntoNotes datasets for FET.  While researchers have considered the benefits of document-level~\cite{ZhangDD18}, and corpus-level~\cite{YaghoobzadehS15} context, here we focus on the sentence-level variant for best contrast  to prior work.
 
Progress in FET has focused primarily on:
\begin{itemize}[leftmargin=*]
    \item \textbf{Better mention representations:} Starting from sparse hand-crafted binary features \cite{LingW12,GillickLGKH14}, the community has moved to distributed representations \cite{YogatamaGL15}, to pre-trained word embeddings with LSTMs \cite{RenHQHJH16,RenHQVJH16,ShimaokaSIR16,AbhishekAA17,ShimaokaSIR17} or CNNs \cite{McCallumVVMR18}, with mention-to-context attention \cite{ZhangDD18}, then to employing pre-trained language models like ELMo \cite{PetersNIGCLZ18} to generate ever better representations \cite{LinJ19}. Our approach builds upon these developments and uses state-of-the-art mention encoders.
   
    \item \textbf{Incorporating the hierarchy:} Most prior works approach the hierarchical typing problem as  \emph{multi-label classification}, without using information in the hierarchical structure, but there are a few exceptions. \citet{RenHQHJH16} proposed an adaptive margin for learning-to-rank so that similar types have a smaller margin; \citet{XuB18} proposed hierarchical loss normalization that penalizes output that violates the hierarchical property; and \citet{McCallumVVMR18} proposed to learn a \emph{subtyping} relation to constrain the type embeddings in the type space. In contrast to these approaches, our coarse-to-fine decoding approach strictly guarantees that the output does not violate the hierarchical property, leading to better performance. HYENA \citep{YosefBHSW12} applied ranking to sibling types in a type hierarchy, but the number of predicted positive types are trained separately with a meta-model, hence does not support neural end-to-end training.
\end{itemize}

Researchers have proposed alternative FET formulations whose types are not formed in a type hierarchy, in particular Ultra-fine entity typing  \cite{ChoiLCZ18,XiongWLYCGW19,OnoeD19}, with a very large set of types derived from phrases mined from a corpus.  FET in KB \cite{JinHLD19} labels mentions to types in a knowledge base with multiple relations, forming a type graph. \citet{DaiDLS19} augments the task with entity linking to KBs. 

\section{Problem Formulation} \label{sec:problem}
  We denote a mention as a tuple $x = (w, l, r)$, where $w = (w_1, \cdots, w_n)$ is the sentential context and the span $[l:r]$ marks a mention of interest in sentence $w$.  That is, the mention of interest is $(w_l, \cdots, w_r)$. Given $x$, a hierarchical entity typing model outputs a set of types $Y$ in the type ontology $\mathcal{Y}$, i.e. $Y \subseteq \mathcal{Y}$. 
  
  Type hierarchies take the form of a forest, where each tree is rooted by a top-level supertype (e.g. {\tt /person}, {\tt /location}, etc.). We add a dummy parent node $\any$ = ``{\tt /}'', the supertype of all entity types, to all the top-level types, effectively transforming a type forest to a type tree. In \autoref{fig:ontologies}, we show 3 type ontologies associated with 3 different datasets (see \autoref{sec:datasets}), with the dummy \any node augmented.
  
  We now introduce some notation for referring to aspects of a type tree. The binary relation ``type $z$ is a subtype of $y$'' is denoted as $z \st y$.\footnote{~Per programming language literature, e.g. the type system $F_{\st}$ that supports subtyping.} The unique parent of a type $y$ in the type tree is denoted $\bar y \in \mathcal{Y}$, where $\bar y$ is undefined for $y=\any$.  The immediate subtypes of $y$ (children nodes) are denoted $\children(y) \subseteq \mathcal{Y}$.  Siblings of $y$, those sharing the same immediate parent, are denoted $\sibling(y) \subseteq \mathcal{Y}$, where $y\notin\sibling(y)$.
  
  In the AIDA FET ontology (see \autoref{fig:ontologies}), the maximum depth of the tree is $L = 3$, and each mention can only be typed with at most 1 type from each level. We term this scenario  \emph{\textbf{single-path}} typing, since there can be only 1 path starting from the root (\any) of the type tree. This is in contrast \emph{\textbf{multi-path}} typing, such as in the BBN dataset, where mentions may be labeled with multiple types on the same level of the tree.

  Additionally, in AIDA, there are mentions labeled such as as {\tt /per/police/<unspecified>}. In FIGER, we find instances with labeled type {\tt /person} but not any further subtype. What does it mean when a mention $x$ is labeled with a \textbf{\emph{partial type path}}, i.e., a type $y$ but none of the subtypes $z <: y$? We consider two interpretations:
  \begin{itemize}[leftmargin=*]
    \item \textbf{Exclusive:}  $x$ is of type $y$, but $x$ is not of any type $z <: y$.
    \item \textbf{Undefined:}  $x$ is of type $y$, but whether it is an instance of some $z <: y$ is unknown.
  \end{itemize}
  
  We devise different strategies to deal with these two conditions. Under the \textbf{\emph{exclusive}} case, we add a dummy \other node to every intermediate branch node in the type tree. For any mention $x$ labeled with type $y$ but none of the subtypes $z <: y$, we add this additional label ``$y$\texttt{/\other}'' to the labels of $x$ (see \autoref{fig:ontologies}: AIDA). For example, if we interpret a partial type path {\tt /person} in FIGER as \emph{exclusive}, we add another type {\tt /person/\other} to that instance. Under the \textbf{\emph{undefined}} case, we do not modify the labels in the dataset. We will see this can make a significant difference depending on the way a specific dataset is annotated.
  
\section{Model}
  \subsection{Mention Representation}
    Hidden representations for entity mentions in sentence $w$ are generated by leveraging recent advances in language model pre-training, e.g. ELMo \cite{PetersNIGCLZ18}.\footnote{~\citet{LinJ19} found that ELMo performs better than BERT \cite{DevlinCLT19} for FET. Our internal experiments also confirm this finding. We hypothesize that this is due to the richer character-level information contained in lower-level ELMo representations that are useful for FET.} The ELMo representation for each token $w_i$ is denoted as $\mathbf{w}_i \in \mathbb{R}^{d_w}$. Dropout is applied with probability $p_{\rm D}$ to the ELMo vectors.

    Our mention encoder largely follows \citet{LinJ19}. First a mention representation is derived using the representations of the words in the mention. We apply a max pooling layer atop the mention after a linear transformation:
    \begin{equation} \label{eq:mention}
        \mathbf{m} = \mathrm{MaxPool} (\mathbf{T}\mathbf{w}_l, \cdots,\mathbf{T} \mathbf{w}_r) \in \mathbb{R}^{d_w} \,.
    \end{equation}
    
    
    Then we employ mention-to-context attention first described in \citet{ZhangDD18} and later employed by \citet{LinJ19}: a context vector $\mathbf{c}$ is generated by attending the sentence with a query vector derived from the mention vector $\mathbf{m}$. We use the multiplicative attention of \newcite{LuongPM15}:
    \begin{align}
        a_{i} &\propto \exp(\mathbf{m}^{\rm T} \mathbf{Q} \mathbf{w}_i) \\
        \mathbf{c} &= \sum_{i=1}^{N} a_i \mathbf{w}_i \in \mathbb{R}^{d_w}
    \end{align}
    
    The final representation for an entity mention is generated via concatenation of the mention and context vector: $[\mathbf{m} ~;~ \mathbf{c}] \in \mathbb{R}^{2d_w}$.
    
  \subsection{Type Scorer}
    We learn a type embedding $\mathbf{y} \in \mathbb{R}^{d_t}$ for each type $y \in \mathcal{Y}$. To score an instance with representation  $[\mathbf{m} ~;~ \mathbf{c}]$, we  pass it through a 2-layer feed-forward network that maps into the same space as the type space $\mathbb{R}^{d_t}$, with $\tanh$ as the nonlinearity.  The final score is an inner product between the transformed feature vector and the type embedding:
    \begin{equation}\label{eq:scorer}
        F(x, y) = \mathrm{FFNN}([\mathbf{m} ~;~ \mathbf{c}]) \cdot \mathbf{y} \,.
    \end{equation}
  
  \subsection{Hierarchical Learning-to-Rank}
    
    We introduce our novel hierarchical learning-to-rank loss that (1) allows for natural multi-label classification and (2) takes the hierarchical ontology into account.
    
    We start with a multi-class hinge loss that ranks positive types above negative types \cite{WestonW99}:
    \begin{equation}\label{eq:flat-svm}
        J_{\rm flat}(x, Y) = \sum_{y \in Y} \sum_{y\pr \notin Y} [\xi - F(x, y) + F(x, y\pr)]_+
    \end{equation}
    where $[x]_+ = \max\{0, x\}$. This is actually learning-to-rank with a ranking SVM \cite{Joachims02}:  the model learns to rank the positive types $y \in Y$ higher than those negative types $y\pr \notin Y$, by imposing a margin $\xi$ between $y$ and $y\pr$: type $y$ should rank higher than $y\pr$ by $\xi$. Note that in \autoref{eq:flat-svm}, since it is a linear SVM, the margin hyperparameter $\xi$ could be just set as 1 (the type embeddings are linearly scalable), and we rely on $L_2$ regularization to constrain the type embeddings.
    
    \paragraph{Multi-level Margins} However, this method considers all candidate types to be \emph{flat} instead of hierarchical --- all types are given the same treatment without any prior on their relative position in the type hierarchy. Intuitively, coarser types (higher in the hierarchy) should be easier to determine (e.g. {\tt /person} vs {\tt /location} should be fairly easy for the model), but fine-grained types (e.g. {\tt /person/artist/singer}) are harder. 
    
    We encode this intuition by \textbf{(i)} learning to rank types \emph{only} on the same level in the type tree; \textbf{(ii)} setting different margin parameters for the ranking model with respect to different levels:
    \begin{equation} \label{eq:hier1}
        \sum_{y \in Y} \sum_{y\pr \in \sibling(y) \setminus Y} [\xi_{\level(y)} - F(x,y) + F(x,y\pr)]_+
    \end{equation}
    
    Here $\level(y)$ is the level of the type $y$: for example, $\level(\texttt{/location}) = 1$, and $\level(\texttt{/person/artist/singer}) = 3$. In \autoref{eq:hier1}, each positive type $y$ is only compared against its negative siblings $\sibling(y) \setminus Y$, and the margin hyperparameter is set to be $\xi_{\level(y)}$, i.e., a margin dependent on which level $y$ is in the tree. Intuitively, we should set $\xi_1 > \xi_2 > \xi_3$ since our model should be able to learn a larger margin between easier pairs: we show that this is superior than using a single margin in our experiments. 
    
    Analogous to the reasoning that in \autoref{eq:flat-svm} the margin $\xi$ can just be 1, only the relative ratios between $\xi$'s are important. For simplicity,\footnote{~We did hyperparameter search on these margin hyperparameters and found  that \autoref{eq:margins} generalized well.} if the  ontology has $L$ levels, we assign
    \begin{equation}  \label{eq:margins}
        \xi_l = L - l + 1 \ .
    \end{equation}
    For example, given an ontology with 3 levels, the margins per level are $(\xi_1, \xi_2, \xi_3) = (3, 2, 1)$. 
    
    \begin{figure}
        \centering
        \includegraphics[width=\linewidth]{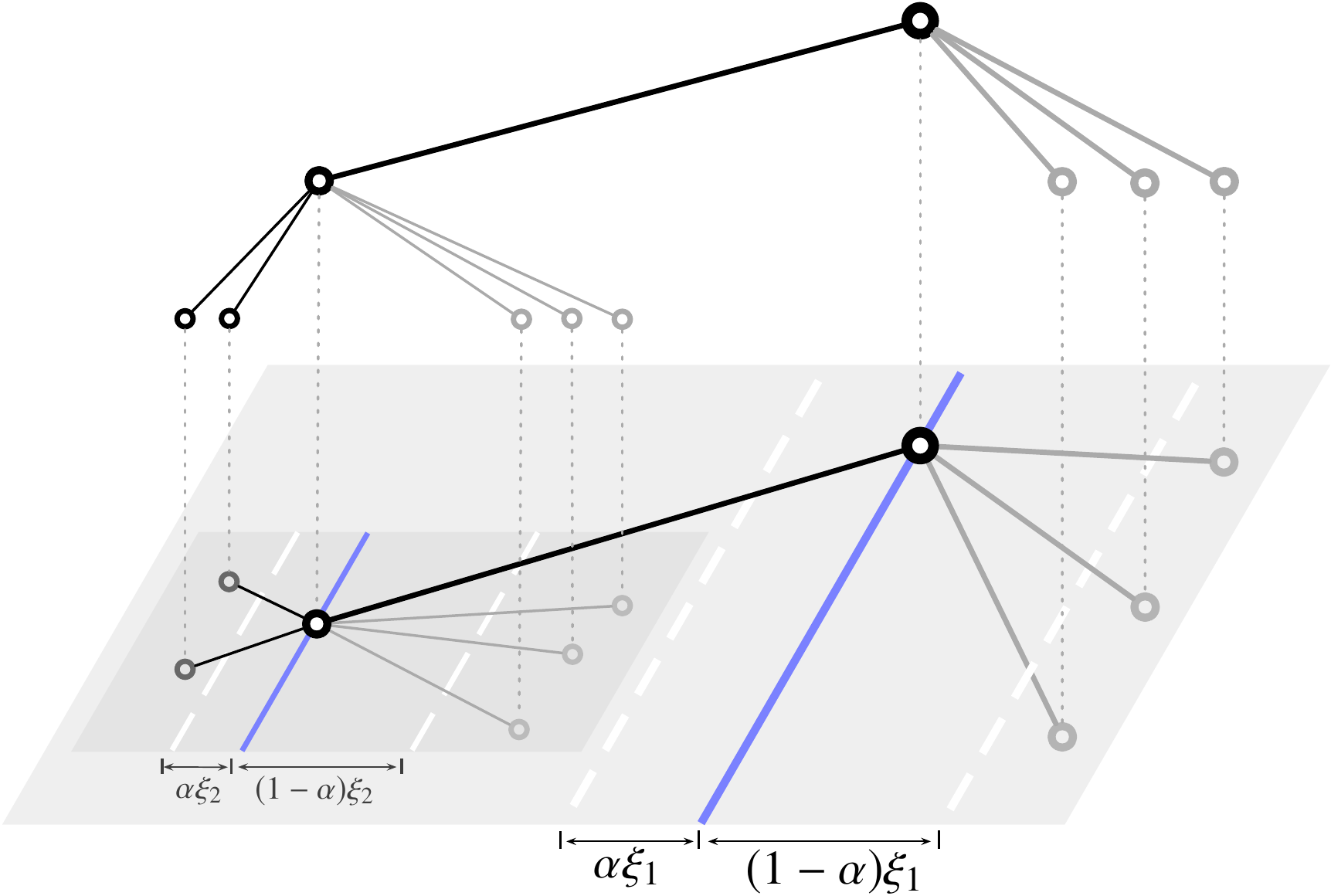}
        \caption{Hierarchical learning-to-rank. Positive type paths are colored black,  negative type paths are colored gray. Each blue line corresponds to a threshold derived from a parent node. Positive types (on the left) are ranked above negative types (on the right).}
        \label{fig:tree-rank}
    \end{figure}
    
    \paragraph{Flexible Threshold} \autoref{eq:hier1} only ranks positive types higher than negative types so that all children types given a parent type are ranked based on their relevance to the entity mention. What should be the threshold between positive and negative types? We could set the threshold to be 0 (approaching the multi-label classification problem as a set of binary classification problem, see \citet{LinJ19}), or tune an adaptive, type-specific threshold for each parent type \cite{ZhangDD18}. Here, we propose a simpler method.
    
    We propose to directly use \emph{the parent node as the threshold}. If a positive type is $y$, we learn the following ranking relation:
    \begin{equation} \label{eq:rank}
        y \succ \parent{y} \succ y\pr, \quad \forall y\pr \in \sibling(t)
    \end{equation}
    where $\succ$ means ``ranks higher than''. For example, a mention has gold type {\tt /person/artist/singer}. Since the parent type {\tt /person/artist} can be considered as a kind of \emph{prior} for all types of artists, the model should learn that the positive type ``singer'' should have a higher confidence than ``artist'', and in turn, higher than other types of artists like ``author'' or ``actor''. Hence the ranker should learn that ``a positive subtype should rank higher than its parent, and its parent should rank higher than its negative children.'' Under this formulation, at decoding time, given parent type $y$, a child subtype $z \st y$ that scores higher than $y$ should be output as a positive label.
    
    We translate the ranking relation in \autoref{eq:rank} into a ranking loss that extends \autoref{eq:hier1}. In \autoref{eq:hier1}, there is an expected margin $\xi$ between positive types and negative types. Since we inserted the parent in the middle, we divide the margin $\xi$ into $\alpha\xi$ and $(1-\alpha)\xi$: $\alpha\xi$ being the margin between positive types and the parent; and $(1 - \alpha)\xi$ is the margin between the parent and the negative types. For a visualization see \autoref{fig:tree-rank}.
    
    The hyperparameter $\alpha \in [0, 1]$ can be used to tune the precision-recall tradeoff when outputting types: the smaller $\alpha$, the smaller the expected margin there is between positive types and the parent. This intuitively increases precision but decreases recall (only very confident types can be output). Vice versa, increasing $\alpha$ decreases precision but increase recall.
    
    Therefore we learn 3 sets of ranking relations from \autoref{eq:rank}: (i) positive types should be scored above parent by $\alpha\xi$; (ii) parent should be scored above any negative sibling types by $(1 - \alpha)\xi$; (iii) positive types should be scored above negative sibling types by $\xi$.  Our final hierarchical ranking loss is formulated as follows.
    \begin{alignat}{10} \label{eq:hier-components}
 J_{y \succ \parent{y}} & = &                                                     & [ &    \alpha &\xi_{\level(y)} & -F(x,         y ) & +&F(x, \parent{y})   & ]_+ \nonumber \\
 J_{\parent{y} \succ y\pr} &=& \displaystyle\sum_{y\pr\in \sibling(y) \setminus Y} & [ & (1-\alpha)&\xi_{\level(y)} & -F(x, \parent{y}) & +&F(x,         y\pr) & ]_+ \nonumber \\
 J_{y \succ y\pr} &=& \displaystyle\sum_{y\pr\in \sibling(y) \setminus Y} & [ &           &\xi_{\level(y)} & -F(x,         y ) & +&F(x,         y\pr) & ]_+ \nonumber
    \end{alignat} \vspace{-0.5cm}
    \begin{equation}
        J_{\rm hier}(x, Y) = \sum_{y \in Y} \left( J_{y \succ \parent{y}} + J_{\parent{y} \succ y\pr} + J_{y \succ y\pr} \right) \\
    \end{equation}
    
  \subsection{Decoding}
  Predicting the types for each entity mention can be performed via iterative searching on the type tree, from the root \any node to coarser types, then to finer-grained types. This ensures that our output does not violate the hierarchical property, i.e., if a subtype is output, its parent must be output.
  
  Given instance $x$ we compute the score $F(x, y)$ for each type $y \in \mathcal{Y}$, the searching process starts with the root node \any of the type tree in the queue. For each type $y$ in the node, a child node $z \st y$ (subtypes) is added to the predicted type set if $F(x, z) > F(x, y)$, corresponding to the ranking relation in \autoref{eq:rank} that the model has learned.
  
  Here we only take the top-$k$ element to add to the queue to prevent from over-generating types. This can also be used to enforce the single-path property  (setting $k$ = 1) if the dataset is single-path. For each level $i$ in the type hierarchy, we limit the branching factor (allowed children) to be $k_i$. The algorithm is listed in \autoref{alg:hiertypedec}, where the function $\textsc{TopK}(S, k, f)$ selects the top-$k$ elements from $S$ with respect to the function $f$.
  
  \begin{algorithm}[t]
  \caption{Decoding for Hierarchical Typing}
  \label{alg:hiertypedec}
  \begin{algorithmic}[1]
    \Procedure{HierTypeDec}{$F(x, \cdot)$} 
      \State $Q \gets \{ \any \}$ \Comment{queue for searching}
      \State $\hat Y \gets \varnothing$ \Comment{set of output types}
      \Repeat
        \State $y \gets \textsc{Dequeue}(Q)$
        \State $\theta \gets F(x, y)$  \Comment{threshold value}
        \State $Z \gets \{ z \in \children(y) \mid F(x, z) > \theta \}$  \par \Comment{all decoded children types}
        \State $Z \gets \textsc{TopK}(Z, k_{\level(y)+1}, F(x, \cdot))$  \par \Comment{pruned by the max branching factors}
        \State $\hat Y \gets \hat Y \cup Z$ 
        \For{$z \in Z$}
          \State $\textsc{Enqueue}(Q, z)$
        \EndFor
       \Until{$Q = \varnothing$} \Comment{queue is empty} \\
      \Return $\hat Y$ \Comment{return all decoded types}
    \EndProcedure
  \end{algorithmic}
\end{algorithm}

  \subsection{Subtyping Relation Constraint} \label{sec:subrelcons}
    Each type $y \in \mathcal{Y}$ in the ontology is assigned a type embedding $\mathbf{y} \in \mathbb{R}^{d_t}$. We notice the binary subtyping relation $\textrm{``}\st\textrm{''} \subseteq \mathcal{Y} \times \mathcal{Y}$ on the types.  \citet{TrouillonWRGB16} proposed the relation embedding method ComplEx that works well with anti-symmetric and transitive relations such as subtyping. It has been  employed in FET before --- in \citet{McCallumVVMR18}, ComplEx is added to the loss to regulate the type embeddings. ComplEx operates in the complex space --- we use the natural isomorphism between real and complex spaces to map the type embedding into complex space (first half of the embedding vector as the real part, and the second half as the imaginary part):
    \begin{align}
        \phi &: \mathbb{R}^{d_t} \to \mathbb{C}^{d_t / 2} \\
        \mathbf{t} &= \left[~ \operatorname{Re}\phi(\mathbf{t}) ~;~ \operatorname{Im}\phi(\mathbf{t})  ~\right] 
    \end{align}
    We learn a single relation embedding $\mathbf{r} \in \mathbb{C}^{d_t / 2}$ for the subtyping relation. Given type $y$ and $z$, the subtyping statement $y\st z$ is modeled using the following scoring function:
    \begin{equation} \label{eq:constraint}
        r(y, z) = \operatorname{Re} \left( \mathbf{r} \cdot \left( \phi(\mathbf{y}) \odot \overline{\phi(\mathbf{z})} 
        \right) \right)
    \end{equation}
    where $\odot$ is element-wise product and $\overline{x}$ is the complex conjugate of $x$. If $y \st z$ then $r(y, z) > 0$; and vice versa, $r(y, z) < 0$ if $y \nless: z$.

    
    \paragraph{Loss}
    Given instance $(x, Y)$, for each positive type $y \in Y$, we learn the following relations:
    \begin{align}
        y & \st \parent{y} \nonumber \\
        y & \nless: y\pr, \quad \forall y\pr \in \sibling(y) \nonumber \\
        y & \nless: y\pr, \quad \forall y\pr \in \sibling(\parent{y})  
    \end{align}
    Translating these relation constraints as a binary classification problem ("is or is not a subtype") under a primal SVM, we get a  hinge loss:
    \begin{align}\label{eq:rel}
         J_{\rm rel}(x, Y)  =& \sum_{y \in Y} \left( \vphantom{\sum_{y \in Y}} [ 1 - r(y, \parent{y}) ]_+ \right.\nonumber \\
                            +& \left. \sum_{y\pr \in  \sibling(y) \cup \sibling(\parent{y})} [ 1 + r(y, y\pr) ]_+ \right) \,.
    \end{align}
    
    This is different from \citet{McCallumVVMR18}, where a binary cross-entropy loss on randomly sampled $(y, y\pr)$ pairs is used. Our experiments showed that the loss in \autoref{eq:rel} performs better than the cross-entropy version,  due to the structure of the training pairs: we use siblings and siblings of parents as negative samples (these are types closer to the positive parent type), hence are training with more competitive negative samples.
  
  \subsection{Training and Validation} 
    Our final loss is a combination of the hierarchical ranking loss and the subtyping relation constraint loss, with $L_2$ regularization:
    \begin{align}
        J_{\rm hier}(x, Y) + \beta J_{\rm rel}(x, Y) + \frac{\lambda}{2} \left\| \boldsymbol{\Theta} \right\|_2^2 \ .
    \end{align}
    
    The AdamW optimizer \cite{LoshchilovH19} is used to train the model, as it is shown to be superior than the original Adam under $L_2$ regularization. Hyperparameters $\alpha$ (ratio of margin above/below threshold), $\beta$ (weight of subtyping relation constraint), and $\lambda$ ($L_2$ regularization coefficient) are tuned.
    
    At validation time, we tune the maximum branching factors for each level $k_1, \cdots, k_L$. These parameters tune the trade-off between the precision and recall for each layer and prevents over-generation (as we observed in some cases). All hyperparameters are tuned so that models achieve maximum micro $F_1$ scores (see \autoref{sec:metrics}).

\section{Experiments}

    \begin{table*}[t!]
        \centering
        \adjustbox{max width=\linewidth}{
        \begin{tabular}{lrrrccc|rrrrr}
          \toprule
             \bf Dataset & \bf  Train & \bf Dev \bf & \bf Test & \bf \# Levels & \bf \# Types & \bf Multi-path? & $\alpha$ & $\beta$ & $\lambda$ & $p_{\rm D}$ & $k_{1, \cdots, L}$  \\
          \midrule
             AIDA      & 2,492  & 558   & 1,383  & 3 & 187 & single-path & 0.1 & 0.3 & 0.1 & 0.5 & (1,1,1) \\
             BBN       & 84,078 & 2,000 & 13,766 & 2 & 56  & multi-path & 0.2 & 0.1 & 0.003 & 0.5 & (2,1)\\
             OntoNotes & 251,039 & 2,202 & 8,963 & 3 & 89  & multi-path & 0.15 & 0.1 & 0.001 & 0.5 & (2,1,1) \\
             FIGER     & 2,000,000 & 10,000 & 563 & 2 & 113 & multi-path &  0.2 & 0.1 & 0.0001 & 0.5 & (2,1) \\
          \bottomrule
        \end{tabular}}
        \caption{Statistics of various datasets and their corresponding hyperparameter settings.}
        \label{tab:stats}
    \end{table*}  
  \subsection{Datasets} \label{sec:datasets}
  
    \paragraph{AIDA}
      The AIDA Phase 1 practice dataset for hierarchical entity typing comprises of 297 documents from \texttt{LDC2019E04} / \texttt{LDC2019E07}, and the evaluation dataset is from \texttt{LDC2019E42} / \texttt{LDC2019E77}. We take only the English part of the data, and use the practice dataset as train/dev, and the evaluation dataset as test. The practice dataset comprises of 3 domains, labeled as \texttt{R103}, \texttt{R105}, and \texttt{R107}. Since the evaluation dataset is out-of-domain, we use the smallest domain \texttt{R105} as dev, and the remaining \texttt{R103} and \texttt{R107} as train. 
      
      The AIDA entity dataset has a 3-level ontology, termed \emph{type}, \emph{subtype}, and \emph{subsubtype}. A mention can only have one label for each level, hence the dataset is \emph{single-path}, thus the branching factors $(k_1, k_2, k_3)$ for the three layers are set to $(1, 1, 1)$.
    
    \paragraph{BBN} \citet{WeischedelB05} labeled a portion of the one million word Penn Treebank corpus of Wall Street Journal texts (\texttt{LDC95T7}) using a two-level hierarchy, resulting in the BBN Pronoun Coreference and Entity Type Corpus. We follow the train/test split by \citet{RenHQVJH16}, and follow the train/dev split by \citet{ZhangDD18}.
    
    \paragraph{OntoNotes} \citet{GillickLGKH14} sampled sentences from the OntoNotes corpus and annotated the entities using 89 types. We follow the train/dev/test data split by \citet{ShimaokaSIR17}. 
    
     \paragraph{FIGER}
      \citet{LingW12} sampled a dataset from Wikipdia articles and news reports. Entity mentions in these texts are mapped to a 113-type ontology derived from Freebase \cite{BollackerEPST08}. Again, we follow the data split by \citet{ShimaokaSIR17}.
  
    The statistics of these datasets and their accompanying ontologies are listed in \autoref{tab:stats}.

  \subsection{Setup}
    
    To best compare to recent prior work, we follow \citet{LinJ19} where the ELMo encodings of words are fixed and not updated. We use all 3 layers of ELMo output, so the initial embedding has dimension $d_w = 3072$. We set the type embedding dimensionality to be $d_t = 1024$. The initial learning rate is $10^{-5}$ and the batch size is 256. 
    
    Hyperparameter choices are tuned on dev sets, and are listed in \autoref{tab:stats}. We employ early stopping: choosing the model that yields the best micro $F_1$ score on dev sets.

    Our models are implemented using AllenNLP \cite{Gardner18AllenNLP}, with implementation for subtyping relation constraints from OpenKE \cite{Han18OpenKE}.
    
  \subsection{Baselines} 
  We compare our approach to major prior work in FET that are capable of \emph{multi-path} entity typing.\footnote{~\citet{ZhangDD18} included document-level information in their best results---for fair comparison, we used their results without document context, as are reported in their ablation tests.} For AIDA, since there are no prior work on this dataset to our knowledge, we also implemented multi-label classification as set of binary classifier models (similar to \citet{LinJ19}) as a baseline, with our mention feature extractor. The results are shown in \autoref{tab:aida-results} as ``Multi-label''.

  \subsection{Metrics} \label{sec:metrics}
  We follow prior work and use strict accuracy (Acc), macro $F_1$ (MaF), and micro $F_1$ (MiF) scores. Given instance $x_i$, we denote the gold type set as $Y_i$ and the predicted type set $\hat Y_i$.  The strict accuracy is the ratio of instances where $Y_i = \hat Y_i$. Macro $F_1$ is the average of all $F_1$ scores between $Y_i$ and $\hat Y_i$ for all instances, whereas micro $F_1$  counts total true positives, false negatives and false positives globally.
  
  We also investigate per-level accuracies on AIDA. The accuracy on level $l$ is the ratio of instances whose predicted type set and gold type set are identical at level $l$. If there is no type output at level $l$, we append with \other to create a dummy type at level $l$: e.g. {\tt /person/\other/\other}. Hence accuracy of the last level (in AIDA, level 3) is equal to the strict accuracy.

  \subsection{Results and Discussions}
    
    All our results are run under the two conditions regarding partial type paths: exclusive or undefined. The result of the AIDA dataset is shown in \autoref{tab:aida-results}. Our model under the exclusive case outperforms a multi-label classification baseline over all metrics.
    
    Of the 187 types specified in the AIDA ontology, the train/dev set only covers 93 types. The test set covers 85 types, of which 63 are seen types. We could perform zero-shot entity typing by initializing a type's embedding using the type name (e.g. {\tt /fac/structure/plaza}) together with its description (e.g. ``\textit{An open urban public space, such as a city square}'') as is designated in the data annotation manual. We leave this as future work.
    
          \begin{table}[H]
        \centering
        \adjustbox{max width=\linewidth}{
        \begin{tabular}{lccccc}
          \toprule
             \bf Approach &  L1 & L2 & L3 & MaF & MiF  \\
          \midrule
            Ours (exclusive)            & \bf 81.6 & 43.1 & \bf 32.0 & \bf 60.6 & \bf 60.0 \\
            Ours (undefined)              & 80.0 & \bf 43.3 & 30.2 & 59.3 & 58.0 \\
          \midrule
            ~~$-$ Subtyping constraints & 80.3 & 40.9 & 29.9 & 59.1 & 58.3 \\
            ~~$-$ Multi-level margins   & 76.9 & 40.2 & 29.8 & 57.4 & 56.9 \\
            Multi-label & 80.5  & 42.1 & 30.7 & 59.7 & 57.9 \\
          \bottomrule
        \end{tabular}}
        \caption{Results on the AIDA dataset.}
        \label{tab:aida-results}
    \end{table}
      
      \begin{table*}[t!]
    \centering
    \adjustbox{max width=\linewidth}{
    \begin{tabular}{lccccccccc}
      \toprule
        \bf Approach   & \multicolumn{3}{c}{\bf BBN} & \multicolumn{3}{c}{\bf OntoNotes} & \multicolumn{3}{c}{\bf FIGER} \\
        \cmidrule(lr){2-4} \cmidrule(lr){5-7} \cmidrule(lr){8-10}
                    & Acc & MaF & MiF        & Acc & MaF & MiF              &   Acc & MaF & MiF    \\
      \midrule 
    \citet{LingW12}  & 46.7 & 67.2 & 61.2 & \multicolumn{3}{c}{$-$ \textsuperscript{\dag}} & 52.3 & 69.9 & 69.3 \\
        \citet{RenHQVJH16} & 49.4 & 68.8 & 64.5 & 51.6 & 67.4 & 62.4 & 49.4 & 68.8 & 64.5  \\
        \citet{RenHQHJH16} & 67.0 & 72.7 & 73.5 & 55.1 & 71.1 & 64.7 & 53.3 & 69.3 & 66.4 \\
        \citet{AbhishekAA17} & 60.4 & 74.1 & 75.7 & 52.2 & 68.5 & 63.3 & 59.0 & 78.0 & 74.9 \\
        \citet{ShimaokaSIR17} & \multicolumn{3}{c}{$-$ \textsuperscript{\dag}} & 51.7 & 71.0 & 64.9 & 59.7 & 79.0 & 75.4 \\ \citet{McCallumVVMR18} &  \multicolumn{3}{c}{$-$ \textsuperscript{\dag}} &  \multicolumn{3}{c}{$-$ \textsuperscript{\dag}}  & 59.7 & 78.3 & 75.4 \\
        \citet{ZhangDD18} & 58.1 & 75.7 & 75.1 & 53.2 & 72.1 & 66.5 & \it \,~60.2\textsuperscript{\ddag} & \it \,~78.7\textsuperscript{\ddag} & \it \,~75.5\textsuperscript{\ddag} \\
        \citet{LinJ19} & 55.9 &  79.3 & 78.1 & \it \,~63.8\textsuperscript{*} & \it \,~82.9\textsuperscript{*} & \it \,~77.3\textsuperscript{*} & 62.9 & \bf 83.0 & 79.8\\
      \midrule \midrule
        Ours (exclusive)  & 48.2 & 63.2 & 61.0  &  58.3 & 72.4 & 67.2  & \bf 69.1 & 82.6 & \bf 80.8 \\
        Ours (undefined)   & \bf 75.2 & \bf 79.7 & \bf 80.5 & \bf 58.7 & \bf 73.0 & \bf 68.1 &  65.5    & 80.5  & 78.1 \\
        \midrule
        ~~$-$ Subtyping constraint & 73.2 & 77.8 & 78.4 & 58.3 & 72.2 & 67.1 & 65.4 & 81.4 & 79.2 \\
        ~~$-$ Multi-level margins & 68.9 & 73.2 & 74.2  & 58.5 & 71.7 & 66.0 & 68.1 & 80.4 & 78.0 \\
      \midrule
        \multicolumn{10}{l}{\small \textsuperscript{\dag}: Not run on the specific dataset; \textsuperscript{*}: Not strictly comparable due to non-standard, much larger training set; } \\
        \multicolumn{10}{l}{\small \textsuperscript{\ddag}: Result has document-level context information, hence not comparable. } \\
      \bottomrule
    \end{tabular}}
    \caption{Results of common FET datasets: BBN, OntoNotes, and FIGER. Numbers in italic are results obtained with various augmentation techniques, either larger data or larger context, hence not directly comparable.}
    \label{tab:results}
  \end{table*}
    
    Results for the BBN, OntoNotes, and FIGER can be found in \autoref{tab:results}. Across 3 datasets, our method produces the state-of-the-art performance on strict accuracy and micro $F_1$ scores, and state-of-the-art or comparable ($\pm 0.5\%$) performance on macro $F_1$ score, as compared to prior models, e.g. \cite{LinJ19}. Especially, our method improves upon the strict accuracy substantially (4\%--8\%) across these datasets, showing our decoder are better at outputting exact correct type sets.
    
    \paragraph{Partial type paths: exclusive or undefined?}
    Interestingly, we found that for AIDA and FIGER, partial type paths should be better considered as \emph{exclusive}, whereas for BBN and OntoNotes, considering them as \emph{undefined} leads to better performance. We hypothesize that this comes from how the data is annotatated---the annotation manual may contain directives as whether to interpret partial type paths as exclusive or undefined, or the data may be non-exhaustively annotated, leading to undefined partial types. We advocate for careful investigation into partial type paths for future experiments and data curation.

    \paragraph{Ablation Studies}
    
  We compare our best model with various components of our model removed, to study the gain from each component. From the best of these two settings (\emph{exclusive} and \emph{undefined}), we report the performance of  (i) removing the subtyping constraint as is described in \autoref{sec:subrelcons}; (ii) substituting the multi-level margins in \autoref{eq:margins} with a ``flat'' margin, i.e., margins on all levels are set to be 1. These results are shown in \autoref{tab:aida-results} and \autoref{tab:results} under our best results, and they show that both multi-level margins and subtyping relation constraints offer orthogonal improvements to our models.

  \paragraph{Error Analysis}
  We identify common patterns of errors, coupled with typical examples: 
  
    \begin{itemize}[leftmargin=*]
  \item {Confusing types}: In BBN, our model outputs {\tt /gpe/city} when the gold type is {\tt /location/region} for ``\emph{... in shipments from the \underline{Valley} of either hardware or software goods.}'' These types are semantically similar, and our model failed to discriminate between these types. 
  
  \item {Incomplete types}: In FIGER, given instance ``\emph{... multi-agency investigation headed by the U.S. \underline{Immigration and Customs Enforcement} 's homeland security investigations unit}'', the gold types are {\tt /government\_agency} and {\tt /organization}, but our model failed to output {\tt /organization}. 
  
  \item {Focusing on only parts of the mention}: In AIDA, given instance ``\emph{... suggested they were the work of \underline{Russian special forces assassins} out to blacken the image of Kievs pro-Western authorities}'', our model outputs {\tt /org/government} whereas the gold type is {\tt /per/militarypersonnel}. Our model focused on the ``Russian special forces'' part, but ignored the ``assassins'' part. Better mention representation is required to correct this, possibly by introducing type-aware mention representation---we leave this as future work.
  \end{itemize}

\section{Conclusions}
We proposed \textbf{(i)} a novel multi-level learning to rank loss function that operates on a type tree, and \textbf{(ii)} an accompanying coarse-to-fine decoder to fully embrace the ontological structure of the types for hierarchical entity typing. Our approach achieved state-of-the-art performance across various datasets, and made substantial improvement (4--8\%) upon strict accuracy. 

Additionally, we advocate for careful investigation into \emph{partial type paths}: their interpretation relies on how the data is annotated, and in turn, influences typing performance. 

\section*{Acknowledgements}
    This research benefited from support by the JHU Human Language Technology Center of Excellence (HLTCOE), and DARPA AIDA. The U.S. Government is authorized to reproduce and distribute reprints for Governmental purposes. The views and conclusions contained in this publication are those of the authors and should not be interpreted as representing official policies or endorsements of DARPA or the U.S. Government.

\bibliographystyle{acl_natbib}
\bibliography{acl2019}

\end{document}